%% file: main.tex
\documentclass[10pt,twocolumn,letterpaper]{article}
\usepackage[T1]{fontenc}
\usepackage[pagenumbers]{cvpr}
\input{preamble}

\definecolor{cvprblue}{rgb}{0.21,0.49,0.74}
\usepackage[breaklinks,colorlinks,allcolors=cvprblue]{hyperref}
\hypersetup{
  pdftitle={Hindsight Self-Distillation: Teaching Robots to Act from the Present},
  pdfauthor={Haisen He, Yechi Liu, Tianle Shi, Haoyun Liu, Chuanzhang Meng, Tiancai Jiao, Yifan Bai, Xinyuan Chang, SongLin Dong, Zhiheng Ma}
}

\title{Hindsight Self-Distillation: Teaching Robots to Act from the Present}
\newcommand{\affilnum}[1]{\textsuperscript{#1}}
\author{%
  Haisen He\affilnum{1,2,3}\quad Yechi Liu\affilnum{4,3}\quad Tianle Shi\affilnum{5}\quad
  Haoyun Liu\affilnum{6,3}\quad Chuanzhang Meng\affilnum{5}\\
  Tiancai Jiao\affilnum{3}\quad Yifan Bai\affilnum{3}\quad Xinyuan Chang\affilnum{3}\quad
  SongLin Dong\affilnum{1}\quad Zhiheng Ma\affilnum{1,3,\textdagger}\\[5pt]
  {\small \affilnum{1}Shenzhen University of Advanced Technology}\quad
  {\small \affilnum{2}Southern University of Science and Technology}\quad
  {\small \affilnum{3}Primotion}\\
  {\small \affilnum{4}New Laboratory of Pattern Recognition (NLPR), Institute of Automation, Chinese Academy of Sciences}\\
  {\small \affilnum{5}Shenzhen University}\quad {\small \affilnum{6}Nanjing University}\\[2pt]
  {\small \textdagger\ Corresponding author}%
}

\begin{document}
\raggedbottom
\maketitle
\input{sec/0_abstract}
\input{sec/1_intro}
\input{sec/2_related_work}
\input{sec/3_method}
\input{sec/4_experiments}
\input{sec/5_conclusion}
\input{sec/6_llm_usage}

{\small
\bibliographystyle{ieeenat_fullname}
\bibliography{main}
}

\input{sec/X_suppl}
\end{document}

%% file: preamble.tex
\usepackage{amsthm}
\newtheorem{proposition}{Proposition}
\usepackage{placeins}

%% file: sec/0_abstract.tex
\begin{abstract}
A robot must choose an action before it can see the outcome. In a recorded demonstration, later camera frames show what happened after the robot moved. Hindsight Self-Distillation (HSD) uses these later frames to teach a policy what to extract from current images. During training, a teacher sees both the current and later frames, while a student sees the current frames. They share a visual encoder and an action predictor. Each branch summarizes its images with a small set of feature vectors. These visual summaries supply all visual input to the action predictor. The teacher learns from demonstrated actions. For each noisy action sequence, the student learns from either the demonstration or the teacher's denoising prediction. It also learns to match the teacher's visual summaries. We analyze which aspects of the teacher's predictions the student can recover from its inputs. At deployment, the student encodes the current images once and reuses its summaries at every denoising step. HSD achieves average success rates of 98.50\% on LIBERO and 92.31\% on RoboTwin 2.0.
\end{abstract}

%% file: sec/1_intro.tex
\section{Introduction}
\label{sec:introduction}

A robot must choose an action before seeing its consequences. A demonstration, by contrast, records both the action and what followed. Subsequent observations reveal how an interaction unfolded, providing hindsight that is unavailable at the moment of action selection. This asymmetry creates a learning opportunity: future outcomes can teach a policy what to extract from the present. Figure~\ref{fig:teaser} shows how a teacher uses these later frames to guide a student that acts from the current observation.

\input{sec/teaser}

Generalist policies draw on vision-language and robot pretraining \citep{zitkovich2023rt2,kim2024openvla,octo2024policy}. Video-based policies use video for planning, for learning predictive features, or for modeling video and actions jointly \citep{du2023unipi,hu2025vpp,zhu2025uwm}. Fast-WAM trains on video and actions and encodes the current observation once at deployment \citep{yuan2026fastwam}. FLARE trains tokens inside the action denoiser to predict future visual features \citep{zheng2025flare}. Being-H0.7 aligns a branch that sees the current context with one that also sees later frames \citep{luo2026being}. We ask how later frames can supervise both the features a policy extracts from current images and the actions it predicts from those features.

In \textbf{Hindsight Self-Distillation (HSD)}, the teacher and student share a video encoder and an action predictor but use separate sets of learned queries. The teacher's queries attend to current and future frames from the demonstration, while the student's queries attend only to current frames. In each branch, the resulting visual summaries supply all visual input to the action predictor, which also receives the robot state and task instruction.

The teacher learns from demonstrated actions and provides two targets for the student within the same policy. For each noisy action sequence, the student learns either from the demonstration or from the teacher's prediction at the same noise level. It also learns to match the teacher's final visual summaries. When a teacher output serves as a target, gradients from that comparison flow through the student branch. When the robot acts, only the student runs. It encodes the current images once per action chunk and reuses the resulting features as it denoises the actions.

With the model weights held fixed, we compare the action distributions of the teacher and student. Their difference has three components: information available only in later frames, information lost when the student summarizes the current observation, and error in the student's approximation. A separate analysis of squared error identifies the part of a teacher target that the student can predict from its inputs.

Our contributions are threefold. \textbf{(i)} We train two sets of visual summaries on demonstrated actions. Each set supplies all visual input for action prediction in its branch. \textbf{(ii)} Within this shared policy, the teacher supervises both the student's denoising predictions and its final visual summaries. \textbf{(iii)} We characterize how later frames, compression of the current images, and student approximation each contribute to the difference between the two action distributions. We evaluate HSD on LIBERO, on task variants in LIBERO-Plus, on RoboTwin 2.0, and on six manipulation tasks with a physical robot. We also vary the two transfer losses and the number of summary tokens, and measure inference latency.

%% file: sec/teaser.tex
\begin{figure*}[t]
    \centering
    \includegraphics[width=.91\textwidth]{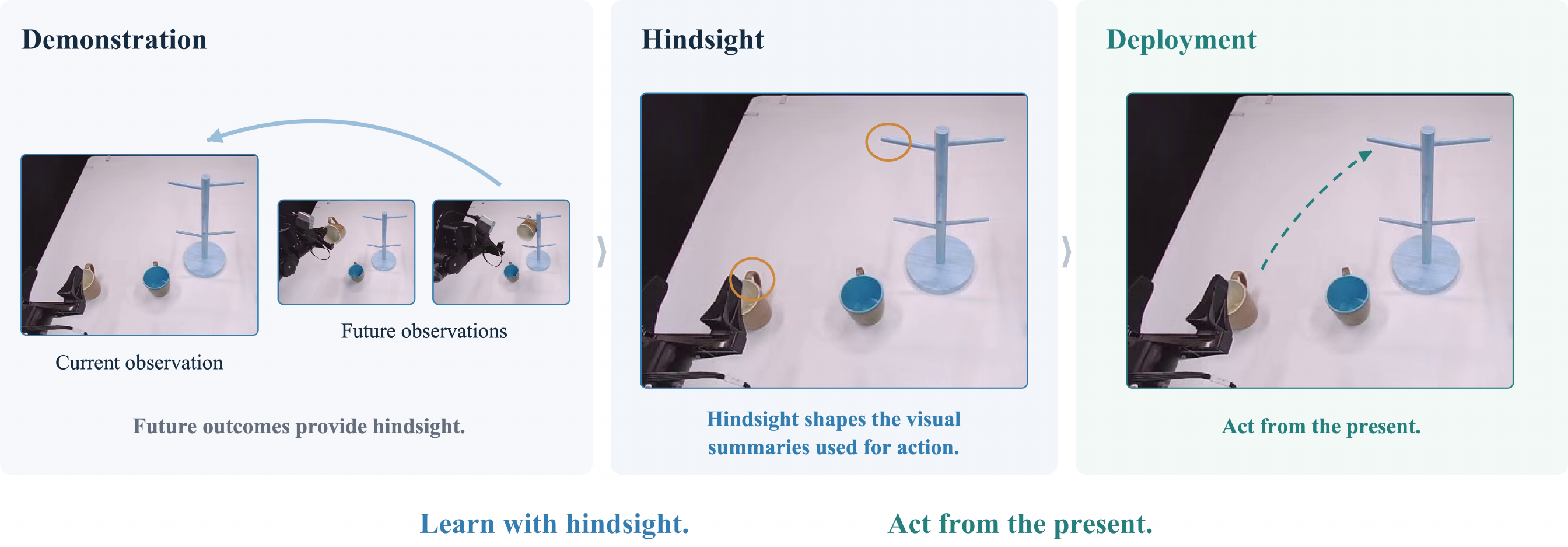}
    \caption{\textbf{Learn with hindsight, act from the present.}
    The demonstration records the scene before and after the robot moves (left).
    During training, the teacher uses the current and later frames to guide what
    the student extracts from the current images (center).
    At deployment, the student chooses its actions from the current images,
    robot state, and instruction (right).
    Circles and arrows illustrate the interaction.}
    \label{fig:teaser}
\end{figure*}

%% file: sec/2_related_work.tex
\section{Related Work}
\label{sec:related_work}
\paragraph{Generalist policies and generative control.}
RT-2 and OpenVLA transfer vision-language pretraining to control, while Octo learns from diverse robot demonstrations \citep{zitkovich2023rt2,kim2024openvla,octo2024policy}. Diffusion Policy and $\pi_0$ generate action chunks through conditional denoising and flow matching \citep{chi2023diffusion,black2025pi0}.

\paragraph{Video models for action generation.}
UniPi converts generated videos into actions; VPP extracts predictive video features \citep{du2023unipi,hu2025vpp}. Unified World Models couples video and action diffusion, whereas UVA supports decoupled decoding \citep{zhu2025uwm,li2025uva}. DreamZero, Cosmos Policy, and LingBot-VA adapt video models for control and world modeling \citep{ye2026dreamzero,kim2026cosmos,li2026lingbot}. Fast-WAM retains video co-training but encodes current observations in one deployment-time pass \citep{yuan2026fastwam}. HSD uses this deployment pattern and learns the visual summaries supplied to the action expert through hindsight supervision.

\paragraph{Future representations for policy learning.}
FLARE trains future tokens in the action denoiser to predict embeddings from an action-aware observation encoder \citep{zheng2025flare}. HSD instead supervises the visual summaries upstream of the denoiser, which reads these summaries at every action step. FRAPPE uses multiple future targets in staged training \citep{zhao2026frappe}. VLA-JEPA pretrains with latent future prediction; JEPA-VLA integrates pretrained video-predictive embeddings into VLA policies \citep{sun2026vlajepa,miao2026jepavla}. LARA jointly optimizes latent-action and VLA models through alignment \citep{liu2026lara}. JEPA-WAM combines structured latent transitions with action generation, and JEPA Policy learns paired action and future-representation targets in a shared Transformer \citep{lin2026jepawam,xu2026jepapolicy}.

\begin{figure*}[tb]
    \centering

    \includegraphics[width=\linewidth]{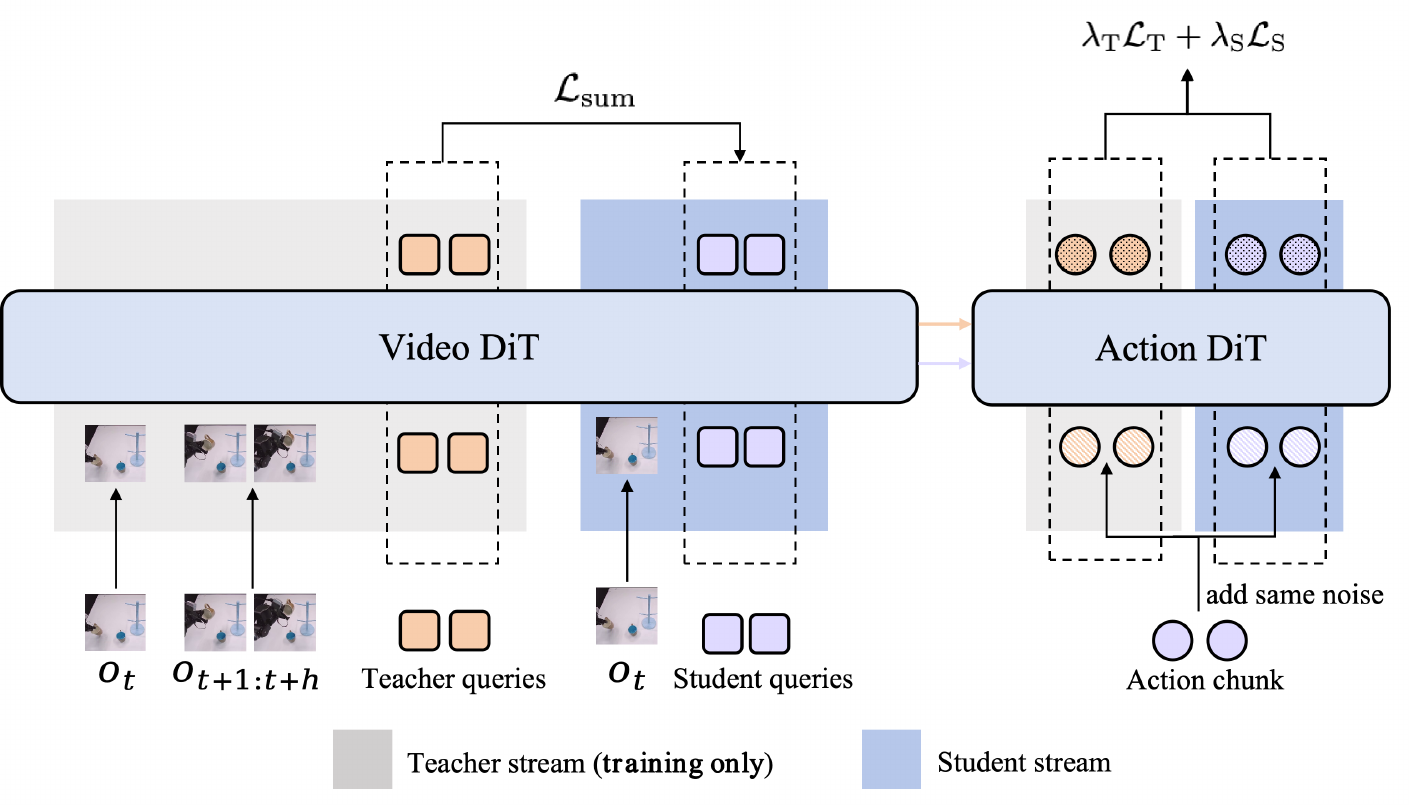}
    \caption{\textbf{Hindsight Self-Distillation.}
    Gray and blue backgrounds denote the training-only teacher and deployable student, which share the Video DiT and Action DiT weights. Teacher queries (orange) summarize current and future observations; student queries (purple) summarize the current observation. Branch-specific summary keys and values provide the sole visual input to each action stream. Both streams receive $A_\tau$ at the same flow time $\tau$. $\mathcal L_{\mathrm T}$ uses ground-truth flow; $\mathcal L_{\mathrm S}$ selects ground-truth flow or a detached teacher prediction per chunk. $\mathcal L_{\mathrm{sum}}$ aligns normalized terminal summaries. Encoders, nonvisual conditions, and auxiliary video co-training are omitted. Deployment retains only the student.}
    \label{fig:method}
\end{figure*}

\paragraph{Privileged supervision and action prediction.}
World Guidance learns future conditions and freezes their encoder before training the policy for deployment \citep{su2026wog}. JALA aligns predictive action embeddings with representations from inverse dynamics, while ForeTime-VLA distills offline targets from future observations into a policy that uses observation history \citep{luo2026jala,ma2026foretime}. DECOWAM distills a bottleneck trained to reconstruct actions and uses it for residual context conditioning \citep{ma2026decowam}. LeapBot-WA uses asymmetric Anchor--Action DiTs and holds the cached features from the current observation fixed during action generation \citep{liu2026leapbot}. Being-H0.7 jointly trains two action branches and aligns their representations across several layers. Both read the shared context, and one also sees future observations \citep{luo2026being}. Privileged Foresight Distillation uses a residual output adapter to transfer corrections derived from future observations \citep{fang2026pfd}. HSD transfers action flow predictions and aligns the visual summaries that supply each branch with all its visual input.

\paragraph{Latent interfaces and self-distillation.}
Perceiver and BLIP-2 compress visual inputs through learnable queries \citep{jaegle2021perceiver,li2023blip2}. Knowledge distillation, FitNets, and Be Your Own Teacher establish prediction transfer, feature supervision, and self-distillation \citep{hinton2015distill,romero2015fitnets,zhang2019byot}; REPA aligns diffusion representations with pretrained visual features \citep{yu2025repa}. HSD trains the queries with action supervision and transfers predictions and summaries from teacher to student.

\paragraph{Privileged learning through a visual interface.}
Privileged learning and generalized distillation formalize the use of information available only during training \citep{vapnik2009lupi,lopezpaz2016generalized}. In partially observed control, a teacher's predictions can depend on evidence the deployed policy cannot observe \citep{warrington2021asymmetric}. We use conditional-information and conditional-expectation identities \citep{cover2006information} to describe what the student can recover from its visual summaries when the model weights are held fixed.

%% file: sec/3_method.tex
\section{Method}
\label{sec:method}

HSD uses two diffusion transformers (Figure~\ref{fig:method}). The video transformer (Video DiT) reads the images and updates the learned queries to produce visual summaries. The action transformer (Action DiT) uses these summaries, the robot state, and the instruction to generate a short sequence of actions, called an action chunk. Teacher and student share both transformers and have separate queries. During training, the teacher also sees later images from the demonstration. Its denoising predictions and final summaries supervise the student. At deployment, the student reuses the summaries from the current images throughout the generation of one action chunk.

\subsection{Setup}
\label{sec:setup}

Each demonstration provides a deployment context $C=(o_t,s_t,\ell)$, future observations $F$, and an action chunk $A\in\mathbb R^{H\times d_a}$. Here $o_t$ contains the current camera views, $s_t$ is proprioception, $\ell$ is the instruction, $H$ is the action horizon, and $d_a$ is the action dimension. Figure~\ref{fig:method} denotes $F$ schematically by $o_{t+1:t+h}$, with observation horizon $h$ distinct from $H$. On LIBERO, a training window spans 32 action steps. The teacher receives the current image from each camera and eight later images from that camera at nominal offsets $4,8,\ldots,32$ action steps. The teacher receives $\widetilde F$, including future corruption and its conditioning metadata. Expectations use the training law $\mathbb P$, covering data sampling, noise, and conditioning augmentation.

We use Fast-WAM's separate video and action experts \citep{yuan2026fastwam}, with the Wan2.2-TI2V-5B video backbone \citep{wan2025models,wan2025wan22}. Teacher and student share both experts; initialization and training settings appear in Section~\ref{sec:exp_setup}.

Action flow matching \citep{lipman2023flow} trains each branch to predict a velocity along the path between a demonstrated action chunk $A$ and Gaussian noise $\epsilon$. Following Fast-WAM, we parameterize the path from data to noise. For $\epsilon\sim\mathcal N(0,I)$ and $\tau\in[0,1]$,
\begin{equation}
 A_\tau=(1-\tau)A+\tau\epsilon,
 \qquad U_\tau=\frac{\mathrm d A_\tau}{\mathrm d\tau}=\epsilon-A.
 \label{eq:flow_path}
\end{equation}
Both branches receive the same $A_\tau$ and $\tau$, with $\tau=0$ denoting data and $\tau=1$ noise. The weighted flow error is
\begin{equation}
 \ell_a(\widehat U,U;\tau)
 =\sum_{k=1}^{Hd_a}\Omega_k(\widehat U_k-U_k)^2,
 \qquad \Omega_k\geq0,
 \label{eq:weighted_action_error}
\end{equation}
where $k$ indexes flattened action coordinates. The weights $\Omega_k$ include timestep weighting, padding masks, and loss normalization; they are shared by both branches and independent of the target selector below. Their dependence on other training variables is implicit.

\subsection{Paired visual summaries}
\label{sec:visual_summaries}

Let $z_0$ and $z_f$ denote the visual tokens for the current and later images, respectively. Each branch starts with $K$ learned query vectors of width $D$. The two sets of queries, $Q^{\mathrm T},Q^{\mathrm S}\in\mathbb R^{K\times D}$, are initialized independently. For branch $b\in\{\mathrm T,\mathrm S\}$, let $Z_0^b=Q^b$ and let $Z_j^b$ be the query states after Video DiT layer $j$, for $j=1,\ldots,L$. These states are the visual summaries. Teacher queries attend to $z_0$ and $z_f$, and student queries attend to $z_0$. Queries within each group also attend to one another. The student's visual timestep is fixed to the clean current observation, independently of the noisy actions and their flow time.

\paragraph{Summary-only visual conditioning.}
At action layer $j$, the Video DiT projects the incoming states $Z_{j-1}^b$ into keys and values $(K_j^b,V_j^b)$ before updating them to $Z_j^b$. These form the complete visual interface
\begin{equation}
 \mathcal K^b=\{(K_j^b,V_j^b)\}_{j=1}^{L},
 \qquad b\in\{\mathrm T,\mathrm S\}.
 \label{eq:summary_interface}
\end{equation}
The shared action expert predicts
\begin{align}
 \widehat U_\tau^{\mathrm T}
 &=v_\theta(A_\tau,\tau,s_t,\ell;
              \mathcal K^{\mathrm T}(C,\widetilde F)),\nonumber\\
 \widehat U_\tau^{\mathrm S}
 &=v_\theta(A_\tau,\tau,s_t,\ell;
              \mathcal K^{\mathrm S}(C)).
 \label{eq:two_branches}
\end{align}
Here $\theta$ includes the shared transformer weights and each branch's learned queries. Each action stream uses the instruction, proprioception, and action self-attention. Its visual input consists entirely of the keys and values derived from its own summaries. Action supervision therefore trains the queries as well as the shared transformers.

\begin{figure}[t]
    \centering
    \includegraphics[width=.83\linewidth]{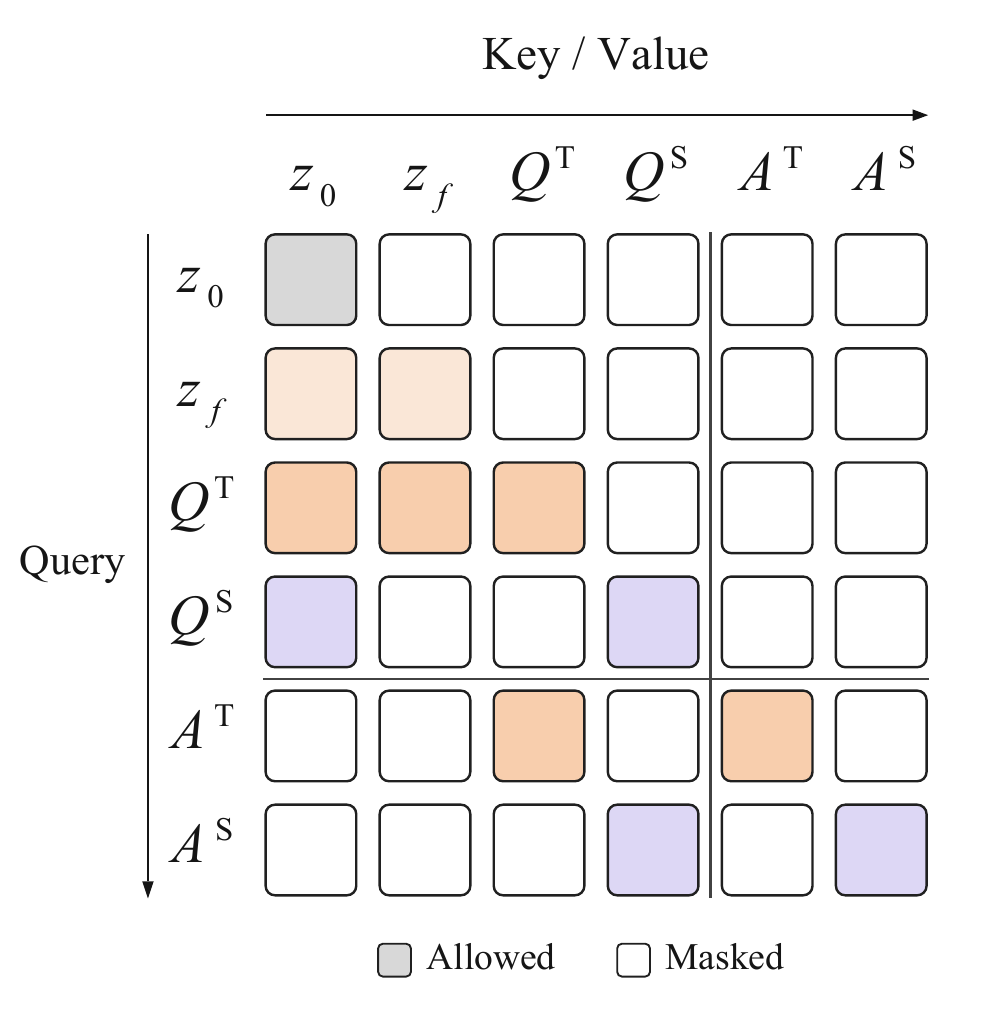}
    \caption{\textbf{Attention within and across branches.} Rows are queries and columns are keys and values. Shaded cells allow attention. Of the two summary groups, only the teacher queries $Q^{\mathrm T}$ read the later visual tokens $z_f$. Each action stream $A^b$ reads its own summaries and action tokens.}
    \label{fig:attention_mask}
\end{figure}

\paragraph{Information flow.}
Both branches run in one forward pass (Figure~\ref{fig:attention_mask}). They share one group of current image tokens, which attend only to one another. The teacher queries can also attend to later frames. The student queries read the current image tokens and one another. At fixed weights and encoder randomness, changing the later frames leaves the student prediction unchanged. During training, the later frames influence the student through the shared weights and the teacher targets. Appendix~\ref{app:implementation} gives the corresponding invariance for the current image encoder and the student prediction.

\subsection{Joint learning and two-level distillation}
\label{sec:objectives}

\paragraph{Learning the teacher.}
Ground-truth flow supervision trains the teacher's action predictor and visual summaries:
\begin{equation}
 \mathcal L_{\mathrm T}
 =\mathbb E\big[\ell_a(\widehat U_\tau^{\mathrm T},U_\tau;\tau)\big].
 \label{eq:teacher_objective}
\end{equation}

\paragraph{Action-level transfer.}
For each example, sample $B\sim\operatorname{Bernoulli}(\rho)$ independently of all other training variables, including data, noise, flow time, future corruption, and loss weights. One selector chooses the target for the entire action chunk:
\begin{equation}
 \begin{aligned}
 Y_{\tau,B}&=(1-B)U_\tau+B\operatorname{sg}(\widehat U_\tau^{\mathrm T}),\\
 \mathcal L_{\mathrm S}&=\mathbb E\big[\ell_a(\widehat U_\tau^{\mathrm S},Y_{\tau,B};\tau)\big],
 \end{aligned}
 \label{eq:student_objective}
\end{equation}
where $\operatorname{sg}$ denotes stop-gradient. Sharing $A_\tau,\tau$ ensures that both targets supervise the same denoising state. Averaging over $B$ weights the ground-truth and detached-teacher errors by $1-\rho$ and $\rho$, respectively.

\paragraph{Summary-level transfer.}
We align the terminal states before the video prediction head:
\begin{equation}
 \mathcal L_{\mathrm{sum}}
 =\frac{1}{KD}\mathbb E\left[
 \left\|\operatorname{LN}_0(Z_L^{\mathrm S})
 -\operatorname{sg}\!\left(\operatorname{LN}_0(Z_L^{\mathrm T})\right)
 \right\|_F^2\right].
 \label{eq:summary_objective}
\end{equation}
Here $\operatorname{LN}_0$ is non-affine LayerNorm over each token's features. Tokens are paired by index, without a learned projector or assignment. This loss aligns terminal representations, while action conditioning uses the layer-wise interface $\mathcal K^b$.

\paragraph{Overall objective.}
An auxiliary noisy-video stream retains the backbone flow loss $\mathcal L_{\mathrm{vid}}$, excluding the conditioned first frame. It has no forward path into student summaries. The total loss is
\begin{equation}
 \mathcal L=\lambda_{\mathrm{vid}}\mathcal L_{\mathrm{vid}}
 +\lambda_{\mathrm T}\mathcal L_{\mathrm T}
 +\lambda_{\mathrm S}\mathcal L_{\mathrm S}
 +\lambda_{\mathrm{sum}}\mathcal L_{\mathrm{sum}}.
 \label{eq:total_objective}
\end{equation}
Teacher targets are detached in both transfer losses. Ground-truth teacher supervision updates its queries and the shared experts; student losses update student queries and shared experts. Thus each transfer gradient follows the student path, while shared-weight updates continue to change both branches.

\subsection{Deployment}
\label{sec:deployment}

At each replanning step, one Video DiT pass constructs $\mathcal K^{\mathrm S}(C)$ from the current observation and student queries. The visual timestep is fixed for the current images, and the summaries are independent of the action tokens and flow time. The action expert therefore reuses the same keys and values as it integrates
\begin{equation}
 \frac{\mathrm d A_\tau}{\mathrm d\tau}
 =v_\theta(A_\tau,\tau,s_t,\ell;\mathcal K^{\mathrm S}(C)).
 \label{eq:deployment_flow}
\end{equation}
Integration starts at $A_1\sim\mathcal N(0,I)$ and runs from $\tau=1$ to $\tau=0$. Only the action expert runs at each denoising step. The cache is refreshed when $C$ changes; teacher computation, future inputs, and auxiliary video denoising are removed.

\subsection{What hindsight transfers}
\label{sec:prediction_boundary}

We compare the two branches at the same model weights. We hold those weights fixed and condition on the random choices made by the encoder for the current images. Action sampling remains random. The identities below hold for each such choice and can be averaged over the encoder randomness. The student retains the robot state, instruction, and visual summaries through their attention keys and values. Together these form $G=g(C)=(s_t,\ell,\mathcal K^{\mathrm S}(C))$, a deterministic function of the full current context $C$ under this conditioning. Let $q_{\mathrm T}(a\mid C,\widetilde F)$ be the teacher's action distribution and define
\[
 \mathbb Q(C,\widetilde F,A^\dagger)
 =\mathbb P(C,\widetilde F)q_{\mathrm T}(A^\dagger\mid C,\widetilde F),
\]
where $A^\dagger$ is a teacher-sampled action, distinct from the demonstration $A$.

\begin{proposition}[Privileged transfer through a visual interface]
\label{prop:prediction_boundary}
For any student distribution $p_{\mathrm S}(a\mid G)$ with finite expected forward KL,
\begin{align}
 &\mathbb E_{C,\widetilde F}D_{\mathrm{KL}}\!\left(
 q_{\mathrm T}(\cdot\mid C,\widetilde F)\,\|\,p_{\mathrm S}(\cdot\mid G)\right)
 \nonumber\\
 &\quad=I_{\mathbb Q}(A^\dagger;\widetilde F\mid C)
       +I_{\mathbb Q}(A^\dagger;C\mid G)\nonumber\\
 &\qquad\quad+\mathbb E_GD_{\mathrm{KL}}\!\left(
 \mathbb Q(\cdot\mid G)\,\|\,p_{\mathrm S}(\cdot\mid G)\right).
 \label{eq:prediction_boundary}
\end{align}
\end{proposition}

The first term measures how much the teacher's actions depend on later images given the current context $C$. The second measures how much information about those actions is lost when $C$ is reduced to $G$. The third compares the student's action distribution with the distribution of teacher actions conditional on $G$ \citep{cover2006information}. Appendix~\ref{app:theory} gives the proof. Among unrestricted conditional distributions, the optimal student is $\mathbb Q(\cdot\mid G)$. A student with the full context $C$ has optimum $\bar q(a\mid C)=\mathbb E_{\widetilde F\mid C}[q_{\mathrm T}(a\mid C,\widetilde F)]$, which averages over futures consistent with $C$.

For HSD's summary regression, set $R=\operatorname{LN}_0(Z_L^{\mathrm T})$, $r_{\mathrm S}(C)=\operatorname{LN}_0(Z_L^{\mathrm S})$, and $m(C)=\mathbb E_{\mathbb P}[R\mid C]$. For square-integrable targets,
\begin{equation}
 \begin{aligned}
 KD\mathcal L_{\mathrm{sum}}&=\mathbb E_{\mathbb P}\|R-m(C)\|_F^2\\
 &\quad+\mathbb E_{\mathbb P}\|m(C)-r_{\mathrm S}(C)\|_F^2.
 \end{aligned}
 \label{eq:summary_prediction_boundary}
\end{equation}
The first term is target variation unpredictable from $C$; the second measures approximation of the predictable mean. This mean is optimal over unrestricted square-integrable predictors, but a normalized student may not represent it exactly. For action-flow transfer, the regression context also includes $A_\tau$ and $\tau$ (Appendix~\ref{app:flow_projection}). The KL identity characterizes teacher-induced distributions under $\mathbb Q$, whereas HSD optimizes regression under $\mathbb P$.

%% file: sec/4_experiments.tex
\section{Experiments}
\label{sec:experiments}

We evaluate the student policy on LIBERO, LIBERO-Plus, RoboTwin 2.0, and six manipulation tasks on a physical robot. We measure success and inference latency and vary the two transfer objectives and the number of summary tokens.

\subsection{Experimental setup}
\label{sec:exp_setup}

\paragraph{Benchmarks and evaluation.}
LIBERO comprises Spatial, Object, Goal, and LIBERO-10 (Long) \citep{liu2023libero}; we run 50 trials per task across 40 tasks. On RoboTwin 2.0 \citep{chen2025robotwin2}, we follow Fast-WAM and train jointly across tasks using 2,500 demonstrations in clean scenes and 25,000 in randomized scenes. We evaluate each task in both conditions with 100 trials per condition \citep{yuan2026fastwam}. Scores average over tasks within each suite or condition; benchmark averages equally weight the four LIBERO suites or two RoboTwin conditions.

\begin{table}[tb]
\centering

\small\setlength{\tabcolsep}{2pt}
\caption{LIBERO success rates (\%). PT marks additional embodied pretraining. DP, MDT, and $\pi_0$ scores follow \citet{kim2025oft}; MDT uses 2\% language annotations. $\pi_{0.5}$ scores follow \citet{kim2026cosmos}. Published averages retain source rounding. The lower block reports our experiments. Bold marks the highest displayed scores, including ties.}
\label{tab:main_results}
\begin{tabular}{@{}lcccccc@{}}
\toprule
Method & PT & Spatial & Object & Goal & Long & Avg. \\
\midrule
\multicolumn{7}{@{}l}{\textit{Previously reported results}} \\
Diffusion Policy \citep{chi2023diffusion} & No & 78.3 & 92.5 & 68.3 & 50.5 & 72.40 \\
MDT \citep{reuss2024mdt} & No & 78.5 & 87.5 & 73.5 & 64.8 & 76.10 \\
OpenVLA-OFT \citep{kim2025oft} & No & 94.3 & 95.2 & 91.7 & 86.5 & 91.90 \\
$\pi_0$ \citep{black2025pi0} & Yes & 96.8 & 98.8 & 95.8 & 85.2 & 94.20 \\
$\pi_{0.5}$ \citep{black2025pi05} & Yes & 98.8 & 98.2 & 98.0 & 92.4 & 96.90 \\
OpenVLA-OFT \citep{kim2025oft} & Yes & 97.6 & 98.4 & 97.9 & 94.5 & 97.10 \\
Cosmos Policy \citep{kim2026cosmos} & No & 98.1 & \textbf{100.0} & 98.2 & 97.6 & \textbf{98.50} \\
Fast-WAM \citep{yuan2026fastwam} & No & 98.2 & \textbf{100.0} & 97.0 & 95.2 & 97.60 \\
Fast-WAM-Joint \citep{yuan2026fastwam} & No & \textbf{99.6} & 99.4 & 98.2 & 96.8 & \textbf{98.50} \\
Fast-WAM-IDM \citep{yuan2026fastwam} & No & 98.8 & 97.8 & 97.8 & 97.6 & 98.00 \\
\midrule
\multicolumn{7}{@{}l}{\textit{Experiments in this work}} \\
Fast-WAM \citep{yuan2026fastwam} & No & 97.0 & 99.4 & 96.6 & 94.8 & 96.95 \\
Being-H0.7-style & No & 98.2 & 99.4 & 97.2 & 93.4 & 97.05 \\
FLARE-style & No & 97.6 & 99.4 & \textbf{98.6} & 93.6 & 97.30 \\
\textbf{HSD (ours)} & No & 98.6 & 99.2 & 98.4 & \textbf{97.8} & \textbf{98.50} \\
\bottomrule
\end{tabular}
\end{table}

\begin{figure*}[!htb]
\centering

\begin{minipage}[t]{0.49\linewidth}
\vspace{0pt}
\centering
\includegraphics[width=\linewidth]{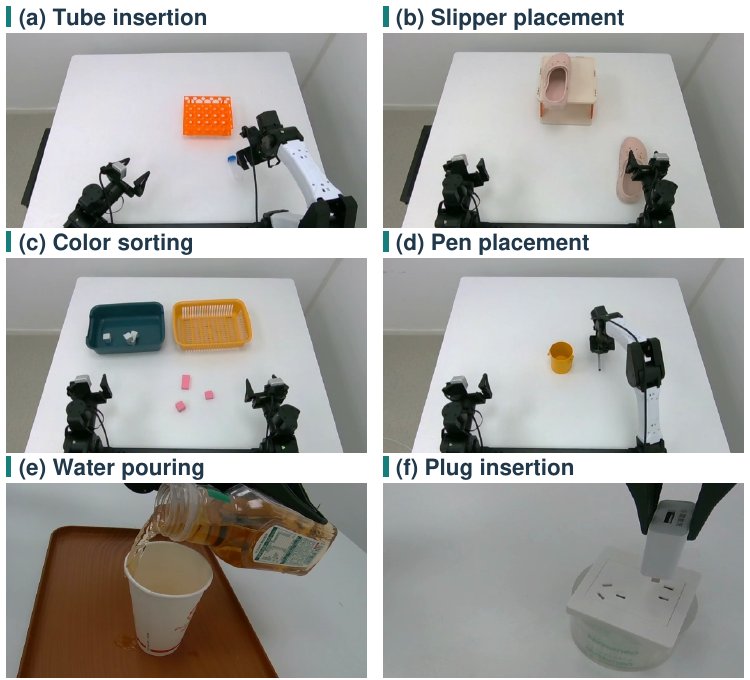}
\caption{\textbf{Six manipulation tasks on the YAM robot.} Demonstration frames show tube insertion, slipper placement, color sorting, pen placement, water pouring, and plug insertion.}
\label{fig:realworld_tasks}
\end{minipage}\hfill
\begin{minipage}[t]{0.49\linewidth}
\vspace{0pt}
\centering
\includegraphics[width=\linewidth]{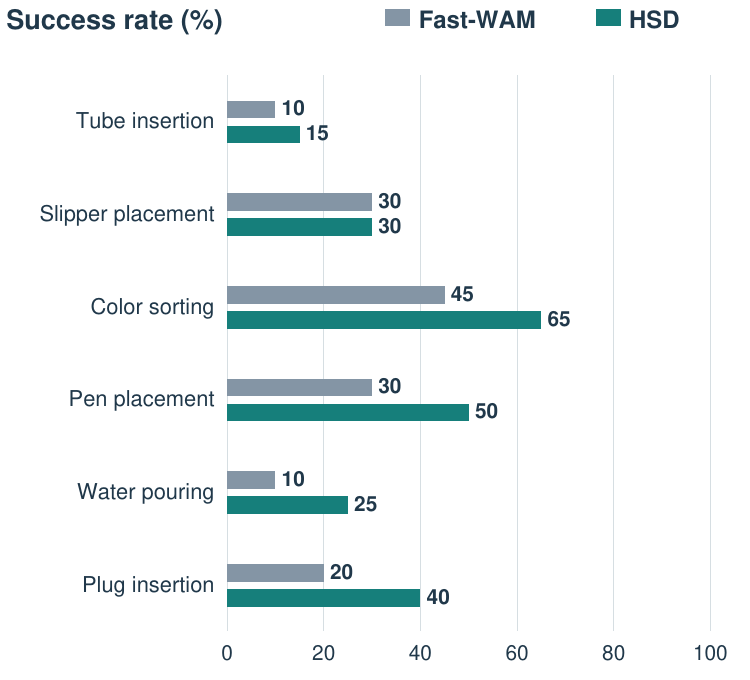}
\caption{\textbf{Real-world success rates.} Each policy is evaluated in 20 trials per task. Mean success across the six tasks is 24.17\% for Fast-WAM and 37.50\% for HSD.}
\label{fig:realworld_results}
\end{minipage}
\end{figure*}

\paragraph{Training recipe.}
We use Fast-WAM's AdamW recipe: learning rate $10^{-4}$, weight decay 0.01, cosine decay, mixed precision, and gradient clipping at 1.0 \citep{yuan2026fastwam}. We train HSD jointly on all four LIBERO suites for 10 epochs and on the full RoboTwin 2.0 training set for 5 epochs. The global batch sizes are 128 and 512, respectively, with 5\% warmup for both.

\paragraph{Model and observations.}
The video backbone is Wan2.2-TI2V-5B, with hidden width $D=3072$. The action expert has hidden width 1024, and both experts use $L=30$ layers. The action expert follows the Fast-WAM initialization procedure, with action horizon $H=32$ and $K=32$ summary tokens per branch. On LIBERO we concatenate two $224\times224$ camera views horizontally. On RoboTwin 2.0, we combine three camera views into a $384\times320$ image. We select the teacher target with probability $\rho=0.5$, set all four loss weights to one, and corrupt the future conditioning with probability 0.1. At deployment, the Video DiT runs once and the action predictor takes ten denoising steps.

\paragraph{Comparison methods.}
Tables~\ref{tab:main_results} and~\ref{tab:robotwin_results} separate published results from our evaluations. PT marks robot or human manipulation pretraining beyond the target benchmark, excluding generic visual, video, and vision-language initialization. We implement Being-H0.7-style and FLARE-style supervision on Fast-WAM. Both use frozen DINOv2-S/14 features \citep{oquab2023dinov2} as future targets and retain Fast-WAM's video loss. In our Being-H0.7 adaptation, learned queries summarize the current view and match query states derived from later frames. The action predictor also reads current image features directly. In our FLARE adaptation, tokens inside the action denoiser predict features of a later image. In HSD, the teacher learns its summaries from demonstrated actions. These summaries supervise the student's summaries, which supply all visual input to its action predictor; the teacher also supervises the action prediction itself.
\subsection{Benchmark results}
\label{sec:exp_main}

\paragraph{LIBERO.}
HSD averages 98.50\% on LIBERO, compared with 96.95\% for our Fast-WAM reproduction, 97.05\% for Being-H0.7-style, and 97.30\% for FLARE-style (Table~\ref{tab:main_results}). The largest gains over both adaptations occur on Long, where HSD succeeds in 97.8\% of trials versus 93.4\% and 93.6\%, respectively. FLARE-style scores highest on Goal (98.6\%). HSD matches the published averages of Cosmos Policy and Fast-WAM-Joint.

\begin{table}[!htb]
\centering
\small\setlength{\tabcolsep}{2.4pt}
\caption{RoboTwin 2.0 success rates (\%) after joint training on clean and randomized scenes. PT marks additional embodied pretraining. Motus and LingBot-VA scores follow their CVPR and RSS versions, respectively. Avg. is the mean of the two displayed conditions. Bold marks column maxima.}
\label{tab:robotwin_results}
\begin{tabular}{@{}lcccc@{}}
\toprule
Method & PT & Clean & Randomized & Avg. \\
\midrule
\multicolumn{5}{@{}l}{\textit{Previously reported results}} \\
$\pi_0$ \citep{black2025pi0} & Yes & 65.92 & 58.40 & 62.16 \\
$\pi_{0.5}$ \citep{black2025pi05} & Yes & 82.74 & 76.76 & 79.75 \\
Motus \citep{bi2025motus} & No & 72.80 & 77.00 & 74.90 \\
Motus \citep{bi2025motus} & Yes & 88.66 & 87.02 & 87.84 \\
LingBot-VA \citep{li2026lingbot} & Yes & 91.99 & 91.11 & 91.55 \\
Fast-WAM \citep{yuan2026fastwam} & No & 91.88 & 91.78 & 91.83 \\
Fast-WAM-Joint \citep{yuan2026fastwam} & No & 90.84 & 90.32 & 90.58 \\
Fast-WAM-IDM \citep{yuan2026fastwam} & No & 91.16 & 91.34 & 91.25 \\
\midrule
\multicolumn{5}{@{}l}{\textit{Our evaluation}} \\
\textbf{HSD (ours)} & No & \textbf{92.48} & \textbf{92.14} & \textbf{92.31} \\
\bottomrule
\end{tabular}
\end{table}

\paragraph{RoboTwin 2.0.}
HSD reaches 92.48\% in Clean and 92.14\% in Randomized scenes, averaging 92.31\% (Table~\ref{tab:robotwin_results}). These exceed the published Fast-WAM scores by 0.60 and 0.36 percentage points, respectively.

\paragraph{Robustness on LIBERO-Plus.}
We evaluate each policy once per variant on LIBERO-Plus \citep{fei2026liberoplus} (Table~\ref{tab:libero_plus}). Overall success is 49.43\% for the released Fast-WAM checkpoint \citep{yuan2026fastwam} and 63.71\% for HSD. The student trained alone reaches 53.48\%, while training with the teacher and no explicit transfer yields 62.83\%. The largest difference between HSD and the variant without explicit transfer is on Long (55.62\% versus 52.76\%). Action transfer alone scores highest on Spatial and Goal.

\begin{table}[!htb]
\centering
\small\setlength{\tabcolsep}{2.3pt}
\caption{Fast-WAM and HSD ablations on LIBERO-Plus. Entries are success rates (\%) over one trial per variant. Overall pools all 10,030 variants. Bold marks the highest score in each column.}
\label{tab:libero_plus}
\begin{tabular}{@{}lrrrrr@{}}
\toprule
Method & Spatial & Object & Goal & Long & Overall \\
\midrule
Fast-WAM (released) & 51.62 & 69.58 & 36.70 & 40.29 & 49.43 \\
\midrule
Student-only & 54.58 & 73.71 & 41.30 & 44.74 & 53.48 \\
No explicit transfer & 68.28 & 78.99 & 51.87 & 52.76 & 62.83 \\
Action only & \textbf{70.98} & 75.97 & \textbf{53.22} & 52.88 & 63.10 \\
Summary only & 61.53 & 78.28 & 45.89 & 51.13 & 59.08 \\
\textbf{HSD (both)} & 70.27 & \textbf{79.27} & 50.37 & \textbf{55.62} & \textbf{63.71} \\
\bottomrule
\end{tabular}
\end{table}

\subsection{Real-world experiments}
\label{sec:realworld}

We evaluate six bimanual tasks on the YAM robot: inserting a tube into a rack, placing a slipper on a shoe rack, sorting blocks by color (gray into the left basket and pink into the right), placing a pen in a container, pouring water from a bottle into a cup, and inserting a plug into a socket (Figure~\ref{fig:realworld_tasks}). For each task, we use 50 demonstrations recording an external view, two wrist views, and both arms' joint and gripper states.

For each task, both methods train for 20,000 steps with a learning rate of $10^{-4}$ and a global batch size of 128. We evaluate each policy in 20 trials per task and score a trial as successful if its final state meets the task goal. Across the six tasks, HSD succeeds in 45 of 120 trials (37.50\%) and Fast-WAM in 29 of 120 (24.17\%), a difference of 13.33 percentage points (Figure~\ref{fig:realworld_results}). Success rises from 10\% to 25\% on water pouring and from 20\% to 40\% on plug insertion. HSD also improves on tube insertion (10\% to 15\%), color sorting (45\% to 65\%), and pen placement (30\% to 50\%). Both policies reach 30\% on slipper placement.

\begin{figure}[!htb]
    \centering

    \includegraphics[width=\linewidth]{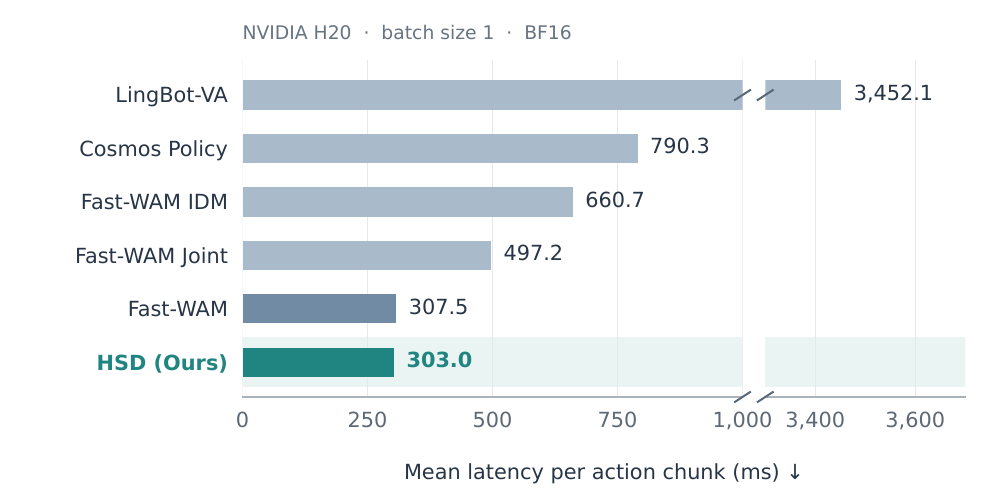}
    \caption{\textbf{Inference latency.} Mean milliseconds per action chunk on one NVIDIA H20 (BF16, batch size 1), using model-specific inference settings. Lower is better.}
    \label{fig:latency}
\end{figure}

\subsection{Component ablations}
\label{sec:exp_ablation}
\begin{table}[tb]
\centering

\small\setlength{\tabcolsep}{2.5pt}
\caption{LIBERO success rates (\%) for ablations of the teacher branch, action-level transfer, and summary-level transfer.}
\label{tab:distill_ablation}
\begin{tabular}{@{}lccccc@{}}
\toprule
Variant & Spatial & Object & Goal & Long & Avg. \\
\midrule
Student-only & \textbf{98.6} & \textbf{99.8} & 97.4 & 92.4 & 97.05 \\
No explicit transfer & \textbf{98.6} & 99.4 & 97.6 & 93.8 & 97.35 \\
Action only & 98.0 & 99.2 & 97.8 & 95.0 & 97.50 \\
Summary only & \textbf{98.6} & 99.4 & 97.0 & 93.4 & 97.10 \\
\textbf{HSD (both)} & \textbf{98.6} & 99.2 & \textbf{98.4} & \textbf{97.8} & \textbf{98.50} \\
\bottomrule
\end{tabular}
\end{table}

\paragraph{Teacher and transfer objectives.}
The student-only baseline trains on demonstrated actions with the auxiliary video loss. The remaining variants retain the teacher and its ground-truth action loss. When enabled, action transfer uses teacher targets with probability $\rho=0.5$, and summary transfer uses weight $\lambda_{\mathrm{sum}}=1$. HSD and student-only average 98.50\% and 97.05\%, respectively, a difference of 1.45 percentage points (Table~\ref{tab:distill_ablation}). The largest difference is on Long (97.8\% vs.\ 92.4\%). Both reach 98.6\% on Spatial, and student-only scores 0.6 points higher on Object. HSD succeeds in 1,970 of 2,000 trials, compared with 1,947 for the variant without explicit transfer, a difference of 1.15 percentage points. Long accounts for 20 of the 23 additional successes (489 versus 469 in 500 trials). Appendix~\ref{app:experiments} gives the evaluation protocol.

\begin{table}[!htb]
\centering
\small\setlength{\tabcolsep}{3.5pt}
\caption{Summary-token count on LIBERO. Both branches use $K$ tokens. Entries are success rates (\%); bold marks the highest score in each column.}
\label{tab:token_budget}
\begin{tabular}{@{}cccccc@{}}
\toprule
$K$ & Spatial & Object & Goal & Long & Avg. \\
\midrule
8 & 97.6 & 98.8 & \textbf{98.8} & 96.4 & 97.90 \\
16 & 98.2 & \textbf{100.0} & 97.8 & 94.8 & 97.70 \\
32 & \textbf{98.6} & 99.2 & 98.4 & \textbf{97.8} & \textbf{98.50} \\
\bottomrule
\end{tabular}
\end{table}

\paragraph{Number of summary tokens.}
Table~\ref{tab:token_budget} compares $K\in\{8,16,32\}$ summary tokens per branch. The corresponding average success rates are 97.90\%, 97.70\%, and 98.50\%. The effect is not monotonic: $K=8$ performs best on Goal, while $K=16$ reaches 100.0\% on Object. We use $K=32$ for its highest overall average and Long success of 97.8\%, compared with 96.4\% for $K=8$ and 94.8\% for $K=16$.

\subsection{Inference latency}
\label{sec:exp_efficiency}

We measure mean latency per action chunk on one NVIDIA H20 with BF16, batch size 1, and cached language conditioning, averaging 100 runs after 30 warm-up runs. HSD takes 303.0\,ms, close to Fast-WAM's 307.5\,ms (Figure~\ref{fig:latency}). Both policies encode the current views once per action chunk and take ten action denoising steps. HSD uses the same cached visual summaries at all ten steps. Fast-WAM-Joint, which generates video and actions together, takes 497.2\,ms per chunk under its inference settings.

%% file: sec/5_conclusion.tex
\FloatBarrier
\section{Conclusion}
\label{sec:conclusion}
HSD uses later frames from a demonstration to train the features a robot uses when it acts. The teacher supplies targets for both the student's visual summaries and its action predictions. At deployment, the student computes its visual summaries from the current images once per action chunk. HSD achieves 98.50\% success on LIBERO and 92.31\% on RoboTwin 2.0. Across six real-world tasks, average success rises from 24.17\% with Fast-WAM to 37.50\% with HSD. The two policies have similar inference latency.

\paragraph{Limitations.}
Hindsight transfer depends on which teacher targets are predictable from the current context. The analysis holds at a fixed snapshot, while closed-loop gains are assessed empirically. Real-world evaluation uses 20 trials per task.

%% file: sec/6_llm_usage.tex
\section*{LLM Usage Statement}
We used OpenAI Codex to assist with writing and formatting the manuscript, checking references, summarizing recorded trial outcomes. The authors take responsibility for the final text, claims, and reported results.

%% file: sec/X_suppl.tex
\clearpage
\maketitlesupplementary

\appendix
\setcounter{equation}{0}
\renewcommand{\theequation}{S\arabic{equation}}
\renewcommand{\theHequation}{supp.\arabic{equation}}
\setcounter{table}{0}
\renewcommand{\thetable}{S\arabic{table}}
\renewcommand{\theHtable}{supp.\arabic{table}}
\section{Analysis of privileged transfer}
\label{app:theory}

\subsection{Conditional-information decomposition}
Fix the model weights and the randomness used to encode the current images. Let $C$ contain the current observation, robot state, and instruction. The student receives $G=g(C)$, which retains the robot state, instruction, and cached visual summaries. Under $\mathbb Q$, we draw $(C,\widetilde F)$ from the training data and augmentation law and sample $A^\dagger$ from the teacher conditioned on both. Assuming regular conditional distributions and finite expected KL divergence as in Proposition~\ref{prop:prediction_boundary} of the main paper, we can write
\begin{align*}
 &\log\frac{q_{\mathrm T}(A^\dagger\mid C,\widetilde F)}
                 {p_{\mathrm S}(A^\dagger\mid G)}\\
 &=\log\frac{q_{\mathrm T}(A^\dagger\mid C,\widetilde F)}
                  {\mathbb Q(A^\dagger\mid C)}
  +\log\frac{\mathbb Q(A^\dagger\mid C)}
                  {\mathbb Q(A^\dagger\mid G)}
  \\&\quad+\log\frac{\mathbb Q(A^\dagger\mid G)}
                  {p_{\mathrm S}(A^\dagger\mid G)}.
\end{align*}
Taking expectation under $\mathbb Q$ yields, respectively,
$I_{\mathbb Q}(A^\dagger;\widetilde F\mid C)$,
$I_{\mathbb Q}(A^\dagger;C\mid G)$, and
$\mathbb E_GD_{\mathrm{KL}}(\mathbb Q(\cdot\mid G)\|p_{\mathrm S}(\cdot\mid G))$.
The conditional-independence identity
$\mathbb Q(A^\dagger\mid C,G)=\mathbb Q(A^\dagger\mid C)$ follows because $G$ is deterministic given $C$. This proves the KL decomposition in the main paper. With an unrestricted distribution family, the final term is minimized by $p_{\mathrm S}(\cdot\mid G)=\mathbb Q(\cdot\mid G)$. With the full context $C$, the optimum averages the teacher distribution over the futures compatible with $C$.

\subsection{Squared-loss summary decomposition}
Let $R$ and $r_{\mathrm S}(C)$ be the teacher and student summaries after normalization in the final layer. With $m(C)=\mathbb E_{\mathbb P}[R\mid C]$,
\[
 R-r_{\mathrm S}(C)=(R-m(C))+(m(C)-r_{\mathrm S}(C)).
\]
The expected cross term vanishes by conditional expectation:
\[
 \mathbb E_{\mathbb P}\langle R-m(C),m(C)-r_{\mathrm S}(C)\rangle_F=0.
\]
Expanding the square proves the summary-loss decomposition in the main paper. The identity holds for the numerical value of the detached teacher target. The conditional mean is the optimum over unrestricted square-integrable predictors. A parameterized, normalized summary stream can have nonzero approximation error because the conditional mean need not lie in its realizable set.

\subsection{Scope of the analysis}
The information decomposition uses the action law $\mathbb Q$ induced by the teacher. The regression decompositions use the training law $\mathbb P$. Both hold for fixed model weights, conditional on the random choices made when encoding the current images. Averaging over that randomness preserves the equalities. The decompositions describe the action distributions and regression targets at those weights. The weights evolve during training, and the policy can visit a different distribution of states in closed-loop execution \citep{ross2011dagger}. The token count specifies how many vectors carry visual features to the action predictor. A Shannon capacity bound would also require assumptions about the information each real-valued vector can carry.

\section{Action-flow regression through the summary interface}
\label{app:flow_projection}

Under the same conditioning on weights and encoder randomness as in Section~\ref{app:theory}, let $T_\tau$ be the numerical value of the detached teacher velocity target. Define
\[
 \begin{aligned}
 X&=(C,A_\tau,\tau), &X_G&=(G,A_\tau,\tau),\\
 m_X&=\mathbb E_{\mathbb P}[T_\tau\mid X], &m_G&=\mathbb E_{\mathbb P}[T_\tau\mid X_G].
 \end{aligned}
\]
For the unweighted squared loss and any square-integrable student predictor $v_{\mathrm S}(X_G)$,
\begin{align}
 &\mathbb E\|T_\tau-v_{\mathrm S}(X_G)\|_F^2\nonumber\\
 &=\mathbb E\|T_\tau-m_X\|_F^2
   +\mathbb E\|m_X-m_G\|_F^2\nonumber\\
 &\quad+\mathbb E\|m_G-v_{\mathrm S}(X_G)\|_F^2.
 \label{eq:flow_projection}
\end{align}
Because $X_G$ is a deterministic function of $X$, the tower property gives
$\mathbb E[m_X\mid X_G]=m_G$. The three summands in
$T_\tau-v_{\mathrm S}=(T_\tau-m_X)+(m_X-m_G)+(m_G-v_{\mathrm S})$
are therefore pairwise orthogonal in $L^2$. The terms describe target variation unexplained by the full regression context, predictable target variation discarded by the interface, and student approximation error within that interface.

\paragraph{Timestep weighting and padding.}
Using the coordinate weights $\Omega_k$ defined in the weighted action loss in the main paper, assume $0<\mathbb E_{\mathbb P}\Omega_k<\infty$ and finite weighted second moments for the target and predictor. Ordinary square integrability under $\mathbb P$ alone is not sufficient if the weights are unbounded. Define the tilted law
\[
 \mathrm d\mathbb P_k^\Omega
 =\frac{\Omega_k}{\mathbb E_{\mathbb P}\Omega_k}\,\mathrm d\mathbb P.
\]
Apply the scalar version of Equation~(\ref{eq:flow_projection}) under $\mathbb P_k^\Omega$, multiply by $\mathbb E_{\mathbb P}\Omega_k$, and sum over $k$. Coordinates with zero expected weight contribute zero. This gives the weighted decomposition without assuming that the padding mask is measurable from deployment information. Its projection targets are weighted conditional means, such as
\[
 m_{G,k}^\Omega
 =\frac{\mathbb E_{\mathbb P}[\Omega_kT_{\tau,k}\mid X_G]}
        {\mathbb E_{\mathbb P}[\Omega_k\mid X_G]},
\]
where the denominator is positive. Values on zero-weight events are immaterial.

\paragraph{Mixture target.}
For the Bernoulli selector independent of all other training variables, the coordinate-wise optimum of the mixed squared loss is
\[
 v_k^*(X_G)=
 \frac{(1-\rho)\mathbb E[\Omega_k U_{\tau,k}\mid X_G]
       +\rho\mathbb E[\Omega_k T_{\tau,k}\mid X_G]}
      {\mathbb E[\Omega_k\mid X_G]}.
\]
For unweighted error this is the corresponding mixture of ordinary conditional means. The teacher term contributes in proportion to $\rho$. The teacher itself is trained against the demonstrated actions.

\paragraph{Random selection and a blended target.}
Let $\bar Y_\tau=(1-\rho)U_\tau+\rho T_\tau$. Conditional on the numerical target values and the selector-independent coordinate weights,
\begin{align}
 &\mathbb E_B\ell_a(\widehat U,Y_{\tau,B};\tau)\nonumber\\
 &=\ell_a(\widehat U,\bar Y_\tau;\tau)
 +\rho(1-\rho)\ell_a(U_\tau,T_\tau;\tau).
 \label{eq:selector_blend}
\end{align}
With the teacher target detached, the last term is independent of the student prediction. For the student prediction, random selection and deterministic blending therefore yield the same expected gradient. Their gradient estimates can have different variances.

\section{Attention mask and summary caching}
\label{app:implementation}

\subsection{Attention mask}
\label{app:block_mask}
Figure~\ref{fig:attention_mask} of the main paper shows the attention mask. Current image tokens attend only to one another. Teacher queries read the current and future image tokens, while student queries read the current tokens. Each set of queries also attends within its own group. Each action branch reads its own summaries and action tokens. The auxiliary video stream shares model weights during training and supplies no inputs to the student action branch.

\subsection{Current-observation encoding}
For fixed encoder randomness, the current image tokens depend only on the current camera views $o_t$,
\[
 z_0=\mathcal E_{\mathrm{cur}}(o_t).
\]
Together with the attention mask and future-independent conditioning, this makes the student's prediction invariant to the future supplied to the teacher at fixed parameters and randomness:
\[
 \widehat U_\tau^{\mathrm S}(C,A_\tau,\tau;\widetilde F_1)
 =\widehat U_\tau^{\mathrm S}(C,A_\tau,\tau;\widetilde F_2).
\]
This follows layer by layer. The current tokens are identical for the two futures at the encoder output. Every subsequent student query reads the same current tokens and previous student states, so its keys and values remain identical. The student action predictor therefore receives the same inputs at every layer.

\subsection{Summary cache}
The student summaries use a fixed visual timestep for the clean current observation. Their attention inputs and modulation are independent of both the noisy action chunk and the action flow time, giving
\[
 \mathcal K^{\mathrm S}(C;A_\tau,\tau)=\mathcal K^{\mathrm S}(C).
\]
At each replanning step, the Video DiT processes the current observation and student queries once. The layer-wise summary keys and values are cached and reused across all action denoising steps. Only the Action DiT receives the changing $A_\tau$ and $\tau$. A new deployment context triggers cache reconstruction.

\section{Experimental details}
\label{app:experiments}

Table~\ref{tab:protocol} summarizes HSD's training and evaluation settings on LIBERO and RoboTwin 2.0. LIBERO training combines all four suites. RoboTwin training combines clean and randomized scenes.

\begin{table*}[tb]
\centering

\caption{HSD training and evaluation settings on LIBERO and RoboTwin 2.0.}
\label{tab:protocol}
\begin{tabular}{lll}
\toprule
Setting & LIBERO & RoboTwin 2.0 \\
\midrule
Video initialization & Wan2.2-TI2V-5B & Wan2.2-TI2V-5B \\
Optimizer & AdamW & AdamW \\
Learning rate & $10^{-4}$ & $10^{-4}$ \\
Training epochs & 10 & 5 \\
Global batch size & 128 & 512 \\
Learning-rate decay & Cosine & Cosine \\
Warmup fraction & 5\% & 5\% \\
Weight decay & 0.01 & 0.01 \\
Gradient clipping & 1.0 & 1.0 \\
Action horizon $H$ & 32 & 32 \\
Summary tokens per branch $K$ & 32 & 32 \\
Video/action hidden width & 3072 / 1024 & 3072 / 1024 \\
Expert layers $L$ & 30 & 30 \\
Teacher-target probability $\rho$ & 0.5 & 0.5 \\
Loss weights $(\lambda_{\mathrm{vid}},\lambda_{\mathrm T},\lambda_{\mathrm S},\lambda_{\mathrm{sum}})$ & $(1,1,1,1)$ & $(1,1,1,1)$ \\
Future corruption probability & 0.1 & 0.1 \\
Action-denoising steps & 10 & 10 \\
Trials per task per condition & 50 & 100 \\
Camera views and final layout & Two, $224\times448$ & Three, $384\times320$ \\
\bottomrule
\end{tabular}
\end{table*}

\paragraph{Fast-WAM adaptations.}
For the Being-H0.7 adaptation \citep{luo2026being}, we use 32 learned prior queries. A resampler pools frozen DINOv2 features from up to eight future frames per camera into 32 posterior tokens. Both action branches also read current image features and fit the demonstrated actions. We align corresponding query states in the last nine layers using mean squared error with gradients to both branches. Our adaptation of FLARE \citep{zheng2025flare} appends 32 future tokens to the action denoiser. A projection at layer 15 predicts frozen DINOv2 features from the two camera views 32 steps ahead using a cosine loss weighted by 0.2. The published FLARE model uses an action-aware target encoder. Both adaptations use current observations at deployment.

\paragraph{Success rates.}
On standard LIBERO, each suite has ten tasks with 50 trials per task. We average task success rates within each suite and give the four suites equal weight. On RoboTwin, we run 100 trials per task in each condition, average over tasks, and report the mean of Clean and Randomized. LIBERO-Plus instead assigns one trial to each variant. Spatial, Object, Goal, and Long contain 2,402, 2,518, 2,591, and 2,519 variants, respectively. Overall success pools all 10,030 trials. The released Fast-WAM checkpoint uses a sampling shift of 5 and HSD uses 1 on LIBERO-Plus.

\paragraph{Transfer ablation on LIBERO.}
Across the 2,000 trials in Table~\ref{tab:distill_ablation} of the main paper, HSD succeeds in 1,970 (98.50\%) and the variant without explicit transfer succeeds in 1,947 (97.35\%), a difference of 1.15 percentage points. Long accounts for 20 of the 23 additional successes (489 versus 469 in 500 trials).

\paragraph{Latency measurement.}
All models use one NVIDIA H20, BF16, batch size 1, and cached language conditioning. We average latency over 100 runs after 30 warm-up runs. Fast-WAM and HSD use two $224\times224$ camera views and ten action-denoising steps, with $H=32$. Fast-WAM-Joint generates video jointly, while Fast-WAM-IDM first performs ten video-denoising steps. LingBot-VA uses two $128\times128$ views, 20 video-denoising steps, and 50 action-denoising steps. Cosmos Policy uses two $224\times224$ views and five sampling steps to jointly predict actions, future states, and values.

%% file: main.bib
@inproceedings{chi2023diffusion,
  title = {{Diffusion Policy: Visuomotor Policy Learning via Action Diffusion}},
  author = {Cheng Chi and Siyuan Feng and Yilun Du and Zhenjia Xu and Eric Cousineau and Benjamin CM Burchfiel and Shuran Song},
  booktitle = {Proceedings of Robotics: Science and Systems},
  year = {2023},
  doi = {10.15607/RSS.2023.XIX.026},
  url = {https://www.roboticsproceedings.org/rss19/p026.html},
}

@inproceedings{black2025pi0,
  title = {{$\pi_0$: A Vision-Language-Action Flow Model for General Robot Control}},
  author = {Kevin Black and Noah Brown and Danny Driess and Adnan Esmail and Michael Robert Equi and Chelsea Finn and Niccolo Fusai and Lachy Groom and Karol Hausman and Brian Ichter and Szymon Jakubczak and Tim Jones and Liyiming Ke and Sergey Levine and Adrian Li-Bell and Mohith Mothukuri and Suraj Nair and Karl Pertsch and Lucy Xiaoyang Shi and Laura Smith and James Tanner and Quan Vuong and Anna Walling and Haohuan Wang and Ury Zhilinsky},
  booktitle = {Proceedings of Robotics: Science and Systems},
  year = {2025},
  doi = {10.15607/RSS.2025.XXI.010},
  url = {https://www.roboticsproceedings.org/rss21/p010.html},
}

@inproceedings{lipman2023flow,
  title = {{Flow Matching for Generative Modeling}},
  author = {Lipman, Yaron and Chen, Ricky T. Q. and Ben-Hamu, Heli and Nickel, Maximilian and Le, Matt},
  booktitle = {International Conference on Learning Representations},
  year = {2023},
  eprint = {2210.02747},
  archivePrefix = {arXiv},
  doi = {10.48550/arXiv.2210.02747},
  url = {https://openreview.net/forum?id=PqvMRDCJT9t},
}

@inproceedings{du2023unipi,
  title = {{Learning Universal Policies via Text-Guided Video Generation}},
  author = {Du, Yilun and Yang, Sherry and Dai, Bo and Dai, Hanjun and Nachum, Ofir and Tenenbaum, Josh and Schuurmans, Dale and Abbeel, Pieter},
  booktitle = {Advances in Neural Information Processing Systems},
  year = {2023},
  volume = {36},
  pages = {9156--9172},
  doi = {10.52202/075280-0403},
  url = {https://proceedings.neurips.cc/paper_files/paper/2023/hash/1d5b9233ad716a43be5c0d3023cb82d0-Abstract-Conference.html},
}

@inproceedings{zhu2025uwm,
  title = {{Unified World Models: Coupling Video and Action Diffusion for Pretraining on Large Robotic Datasets}},
  author = {Chuning Zhu and Raymond Yu and Siyuan Feng and Benjamin Burchfiel and Paarth Shah and Abhishek Gupta},
  booktitle = {Proceedings of Robotics: Science and Systems},
  year = {2025},
  doi = {10.15607/RSS.2025.XXI.015},
  url = {https://www.roboticsproceedings.org/rss21/p015.html},
}

@inproceedings{li2026lingbot,
  title = {{Causal World Modeling for Robot Control}},
  author = {Lin Li and Qihang Zhang and Yiming Luo and Shuai Yang and Ruilin Wang and Luyao Zhang and Mingrui Yu and Zelin Gao and Nan Xue and Boyu Zhou and Xing Zhu and Mingyu Ding and Yujun Shen and Yinghao Xu},
  year = {2026},
  eprint = {2601.21998},
  archivePrefix = {arXiv},
  doi = {10.15607/RSS.2026.XXII.016},
  url = {https://www.roboticsproceedings.org/rss22/p016.html},
  booktitle = {Proceedings of Robotics: Science and Systems},
}

@misc{yuan2026fastwam,
  title = {{Fast-WAM: Do World Action Models Need Test-time Future Imagination?}},
  author = {Yuan, Tianyuan and Dong, Zibin and Liu, Yicheng and Zhao, Hang},
  year = {2026},
  eprint = {2603.16666},
  archivePrefix = {arXiv},
  doi = {10.48550/arXiv.2603.16666},
  url = {https://arxiv.org/abs/2603.16666},
}

@misc{fang2026pfd,
  title = {{Privileged Foresight Distillation: Zero-Cost Future Correction for World Action Models}},
  author = {Fang, Pengcheng and Chen, Hongli and Cai, Xiaohao},
  year = {2026},
  eprint = {2604.25859},
  archivePrefix = {arXiv},
  doi = {10.48550/arXiv.2604.25859},
  url = {https://arxiv.org/abs/2604.25859},
}

@misc{su2026wog,
  title = {{World Guidance: World Modeling in Condition Space for Action Generation}},
  author = {Su, Yue and Chen, Sijin and Shi, Haixin and Liu, Mingyu and Zhang, Zhengshen and Huang, Ningyuan and Zhong, Weiheng and Zhu, Zhengbang and Liu, Yuxiao and Liu, Xihui},
  year = {2026},
  eprint = {2602.22010},
  archivePrefix = {arXiv},
  doi = {10.48550/arXiv.2602.22010},
  url = {https://arxiv.org/abs/2602.22010},
}

@misc{luo2026being,
  title = {{Being-H0.7: A Latent World-Action Model from Egocentric Videos}},
  author = {Luo, Hao and Zhang, Wanpeng and Feng, Yicheng and Zheng, Sipeng and Xu, Haiweng and Xu, Chaoyi and Xi, Ziheng and Fu, Yuhui and Lu, Zongqing},
  year = {2026},
  eprint = {2605.00078},
  archivePrefix = {arXiv},
  doi = {10.48550/arXiv.2605.00078},
  url = {https://arxiv.org/abs/2605.00078},
}

@misc{ma2026foretime,
  title = {{ForeTime-VLA: Causal Future-Token Distillation from a World Action Model for Conveyor-Belt Manipulation}},
  author = {Ma, Siyuan and Zhang, Yutian and Zhang, Boshi and Wu, Qinglian and Zhai, Jiaqi and Wei, Dong and Huang, Xiaojin},
  year = {2026},
  eprint = {2608.20735},
  archivePrefix = {arXiv},
  doi = {10.48550/arXiv.2608.20735},
  url = {https://arxiv.org/abs/2608.20735},
}

@misc{ma2026decowam,
  title = {{DECOWAM: Decoupled Whole-Body World-Action Model for Legged Mobile Manipulation}},
  author = {Ma, Siyuan and Zhang, Boshi and Zhang, Yutian and Wu, Qinglian and Zhai, Jiaqi and Wei, Dong and Yu, Qiaojun},
  year = {2026},
  eprint = {2608.20114},
  archivePrefix = {arXiv},
  doi = {10.48550/arXiv.2608.20114},
  url = {https://arxiv.org/abs/2608.20114},
}

@inproceedings{romero2015fitnets,
  title = {{FitNets: Hints for Thin Deep Nets}},
  author = {Romero, Adriana and Ballas, Nicolas and Kahou, Samira Ebrahimi and Chassang, Antoine and Gatta, Carlo and Bengio, Yoshua},
  booktitle = {International Conference on Learning Representations},
  year = {2015},
  eprint = {1412.6550},
  archivePrefix = {arXiv},
  doi = {10.48550/arXiv.1412.6550},
  url = {https://arxiv.org/abs/1412.6550},
}

@inproceedings{jaegle2021perceiver,
  title = {{Perceiver: General Perception with Iterative Attention}},
  author = {Andrew Jaegle and Felix Gimeno and Andy Brock and Oriol Vinyals and Andrew Zisserman and Joao Carreira},
  booktitle = {Proceedings of the 38th International Conference on Machine Learning},
  year = {2021},
  volume = {139},
  pages = {4651--4664},
  publisher = {PMLR},
  eprint = {2103.03206},
  archivePrefix = {arXiv},
  doi = {10.48550/arXiv.2103.03206},
  url = {https://proceedings.mlr.press/v139/jaegle21a.html},
}

@inproceedings{li2023blip2,
  title = {{BLIP-2: Bootstrapping Language-Image Pre-training with Frozen Image Encoders and Large Language Models}},
  author = {Junnan Li and Dongxu Li and Silvio Savarese and Steven Hoi},
  booktitle = {Proceedings of the 40th International Conference on Machine Learning},
  year = {2023},
  volume = {202},
  pages = {19730--19742},
  publisher = {PMLR},
  eprint = {2301.12597},
  archivePrefix = {arXiv},
  doi = {10.48550/arXiv.2301.12597},
  url = {https://proceedings.mlr.press/v202/li23q.html},
}

@article{vapnik2009lupi,
  title = {{A new learning paradigm: Learning using privileged information}},
  author = {Vapnik, Vladimir and Vashist, Akshay},
  journal = {Neural Networks},
  year = {2009},
  volume = {22},
  number = {5--6},
  pages = {544--557},
  doi = {10.1016/j.neunet.2009.06.042},
  url = {https://doi.org/10.1016/j.neunet.2009.06.042},
}

@inproceedings{lopezpaz2016generalized,
  title = {{Unifying distillation and privileged information}},
  author = {Lopez-Paz, David and Bottou, L\'{e}on and Sch\"{o}lkopf, Bernhard and Vapnik, Vladimir},
  booktitle = {International Conference on Learning Representations},
  year = {2016},
  eprint = {1511.03643},
  archivePrefix = {arXiv},
  doi = {10.48550/arXiv.1511.03643},
  url = {https://arxiv.org/abs/1511.03643},
}

@inproceedings{warrington2021asymmetric,
  title = {{Robust Asymmetric Learning in POMDPs}},
  author = {Andrew Warrington and Jonathan W Lavington and Adam Scibior and Mark Schmidt and Frank Wood},
  booktitle = {Proceedings of the 38th International Conference on Machine Learning},
  year = {2021},
  volume = {139},
  pages = {11013--11023},
  publisher = {PMLR},
  eprint = {2012.15566},
  archivePrefix = {arXiv},
  doi = {10.48550/arXiv.2012.15566},
  url = {https://proceedings.mlr.press/v139/warrington21a.html},
}

@book{cover2006information,
  title = {{Elements of Information Theory}},
  author = {Cover, Thomas M. and Thomas, Joy A.},
  year = {2006},
  edition = {2},
  publisher = {Wiley},
  doi = {10.1002/047174882X},
  url = {https://onlinelibrary.wiley.com/doi/book/10.1002/047174882X},
}

@inproceedings{ross2011dagger,
  title = {{A Reduction of Imitation Learning and Structured Prediction to No-Regret Online Learning}},
  author = {Stephane Ross and Geoffrey Gordon and Drew Bagnell},
  booktitle = {Proceedings of the Fourteenth International Conference on Artificial Intelligence and Statistics},
  year = {2011},
  volume = {15},
  pages = {627--635},
  publisher = {PMLR},
  eprint = {1011.0686},
  archivePrefix = {arXiv},
  doi = {10.48550/arXiv.1011.0686},
  url = {https://proceedings.mlr.press/v15/ross11a.html},
}

@inproceedings{zhang2019byot,
  title = {{Be Your Own Teacher: Improve the Performance of Convolutional Neural Networks via Self Distillation}},
  author = {Zhang, Linfeng and Song, Jiebo and Gao, Anni and Chen, Jingwei and Bao, Chenglong and Ma, Kaisheng},
  booktitle = {Proceedings of the IEEE/CVF International Conference on Computer Vision},
  year = {2019},
  pages = {3712--3721},
  doi = {10.1109/ICCV.2019.00381},
  url = {https://www.computer.org/csdl/proceedings-article/iccv/2019/480300d712/1hVlMcF0InK},
}

@inproceedings{luo2026jala,
  title = {{Joint-Aligned Latent Action: Towards Scalable VLA Pretraining in the Wild}},
  author = {Luo, Hao and Wang, Ye and Zhang, Wanpeng and Yuan, Haoqi and Feng, Yicheng and Xu, Haiweng and Zheng, Sipeng and Lu, Zongqing},
  booktitle = {Proceedings of the IEEE/CVF Conference on Computer Vision and Pattern Recognition (CVPR)},
  year = {2026},
  pages = {35047--35058},
  eprint = {2602.21736},
  archivePrefix = {arXiv},
  doi = {10.48550/arXiv.2602.21736},
  url = {https://openaccess.thecvf.com/content/CVPR2026/html/Luo_Joint-Aligned_Latent_Action_Towards_Scalable_VLA_Pretraining_in_the_Wild_CVPR_2026_paper.html},
}

@misc{wan2025models,
  title = {{Wan: Open and Advanced Large-Scale Video Generative Models}},
  author = {{Wan Team} and Wang, Ang and Ai, Baole and Wen, Bin and Mao, Chaojie and Xie, Chen-Wei and Chen, Di and Yu, Feiwu and Zhao, Haiming and Yang, Jianxiao and Zeng, Jianyuan and Wang, Jiayu and Zhang, Jingfeng and Zhou, Jingren and Wang, Jinkai and Chen, Jixuan and Zhu, Kai and Zhao, Kang and Yan, Keyu and Huang, Lianghua and Feng, Mengyang and Zhang, Ningyi and Li, Pandeng and Wu, Pingyu and Chu, Ruihang and Feng, Ruili and Zhang, Shiwei and Sun, Siyang and Fang, Tao and Wang, Tianxing and Gui, Tianyi and Weng, Tingyu and Shen, Tong and Lin, Wei and Wang, Wei and Wang, Wei and Zhou, Wenmeng and Wang, Wente and Shen, Wenting and Yu, Wenyuan and Shi, Xianzhong and Huang, Xiaoming and Xu, Xin and Kou, Yan and Lv, Yangyu and Li, Yifei and Liu, Yijing and Wang, Yiming and Zhang, Yingya and Huang, Yitong and Li, Yong and Wu, You and Liu, Yu and Pan, Yulin and Zheng, Yun and Hong, Yuntao and Shi, Yupeng and Feng, Yutong and Jiang, Zeyinzi and Han, Zhen and Wu, Zhi-Fan and Liu, Ziyu},
  year = {2025},
  eprint = {2503.20314},
  archivePrefix = {arXiv},
  doi = {10.48550/arXiv.2503.20314},
  url = {https://arxiv.org/abs/2503.20314},
}

@misc{wan2025wan22,
  title = {{Wan2.2-TI2V-5B}},
  author = {{Wan Team}},
  year = {2025},
  url = {https://huggingface.co/Wan-AI/Wan2.2-TI2V-5B},
}

@inproceedings{liu2023libero,
  title = {{LIBERO: Benchmarking Knowledge Transfer for Lifelong Robot Learning}},
  author = {Liu, Bo and Zhu, Yifeng and Gao, Chongkai and Feng, Yihao and Liu, Qiang and Zhu, Yuke and Stone, Peter},
  booktitle = {Advances in Neural Information Processing Systems},
  year = {2023},
  volume = {36},
  pages = {44776--44791},
  doi = {10.52202/075280-1939},
  url = {https://proceedings.neurips.cc/paper_files/paper/2023/hash/8c3c666820ea055a77726d66fc7d447f-Abstract-Datasets_and_Benchmarks.html},
  doi_type = {Published version},
}

@inproceedings{bi2025motus,
  title = {{Motus: A Unified Latent Action World Model}},
  author = {Bi, Hongzhe and Tan, Hengkai and Xie, Shenghao and Wang, Zeyuan and Huang, Shuhe and Liu, Haitian and Zhao, Ruowen and Feng, Yao and Xiang, Chendong and Rong, Yinze and Zhao, Hongyan and Liu, Hanyu and Su, Zhizhong and Ma, Lei and Su, Hang and Zhu, Jun},
  booktitle = {Proceedings of the IEEE/CVF Conference on Computer Vision and Pattern Recognition (CVPR)},
  year = {2026},
  month = {June},
  pages = {35101--35113},
  eprint = {2512.13030},
  archivePrefix = {arXiv},
  doi = {10.48550/arXiv.2512.13030},
  url = {https://openaccess.thecvf.com/content/CVPR2026/html/Bi_Motus_A_Unified_Latent_Action_World_Model_CVPR_2026_paper.html},
  doi_type = {arXiv preprint},
}

@misc{chen2025robotwin2,
  title = {{RoboTwin 2.0: A Scalable Data Generator and Benchmark with Strong Domain Randomization for Robust Bimanual Robotic Manipulation}},
  author = {Chen, Tianxing and Chen, Zanxin and Chen, Baijun and Cai, Zijian and Liu, Yibin and Li, Zixuan and Liang, Qiwei and Lin, Xianliang and Ge, Yiheng and Gu, Zhenyu and Deng, Weiliang and Guo, Yubin and Nian, Tian and Xie, Xuanbing and Chen, Qiangyu and Su, Kailun and Xu, Tianling and Liu, Guodong and Hu, Mengkang and Gao, Huan-ang and Wang, Kaixuan and Liang, Zhixuan and Qin, Yusen and Yang, Xiaokang and Luo, Ping and Mu, Yao},
  year = {2025},
  eprint = {2506.18088},
  archivePrefix = {arXiv},
  doi = {10.48550/arXiv.2506.18088},
  url = {https://arxiv.org/abs/2506.18088},
  doi_type = {arXiv preprint},
}

@misc{hinton2015distill,
  title = {{Distilling the Knowledge in a Neural Network}},
  author = {Hinton, Geoffrey and Vinyals, Oriol and Dean, Jeff},
  year = {2015},
  eprint = {1503.02531},
  archivePrefix = {arXiv},
  doi = {10.48550/arXiv.1503.02531},
  url = {https://arxiv.org/abs/1503.02531},
  doi_type = {arXiv preprint},
}

@inproceedings{hu2025vpp,
  title = {{Video Prediction Policy: A Generalist Robot Policy with Predictive Visual Representations}},
  author = {Yucheng Hu and Yanjiang Guo and Pengchao Wang and Xiaoyu Chen and Yen-Jen Wang and Jianke Zhang and Koushil Sreenath and Chaochao Lu and Jianyu Chen},
  booktitle = {Proceedings of the 42nd International Conference on Machine Learning},
  year = {2025},
  volume = {267},
  pages = {24328--24346},
  publisher = {PMLR},
  eprint = {2412.14803},
  archivePrefix = {arXiv},
  doi = {10.48550/arXiv.2412.14803},
  url = {https://proceedings.mlr.press/v267/hu25g.html},
  doi_type = {arXiv preprint},
}

@inproceedings{kim2024openvla,
  title = {{OpenVLA: An Open-Source Vision-Language-Action Model}},
  author = {Moo Jin Kim and Karl Pertsch and Siddharth Karamcheti and Ted Xiao and Ashwin Balakrishna and Suraj Nair and Rafael Rafailov and Ethan P Foster and Pannag R Sanketi and Quan Vuong and Thomas Kollar and Benjamin Burchfiel and Russ Tedrake and Dorsa Sadigh and Sergey Levine and Percy Liang and Chelsea Finn},
  booktitle = {Proceedings of The 8th Conference on Robot Learning},
  year = {2025},
  volume = {270},
  pages = {2679--2713},
  publisher = {PMLR},
  eprint = {2406.09246},
  archivePrefix = {arXiv},
  doi = {10.48550/arXiv.2406.09246},
  url = {https://proceedings.mlr.press/v270/kim25c.html},
  doi_type = {arXiv preprint},
}

@inproceedings{kim2025oft,
  title = {{Fine-Tuning Vision-Language-Action Models: Optimizing Speed and Success}},
  author = {Moo Jin Kim and Chelsea Finn and Percy Liang},
  booktitle = {Proceedings of Robotics: Science and Systems},
  year = {2025},
  doi = {10.15607/RSS.2025.XXI.017},
  url = {https://www.roboticsproceedings.org/rss21/p017.html},
  doi_type = {Published version},
}

@misc{kim2026cosmos,
  title = {{Cosmos Policy: Fine-Tuning Video Models for Visuomotor Control and Planning}},
  author = {Kim, Moo Jin and Gao, Yihuai and Lin, Tsung-Yi and Lin, Yen-Chen and Ge, Yunhao and Lam, Grace and Liang, Percy and Song, Shuran and Liu, Ming-Yu and Finn, Chelsea and Gu, Jinwei},
  year = {2026},
  eprint = {2601.16163},
  archivePrefix = {arXiv},
  doi = {10.48550/arXiv.2601.16163},
  url = {https://arxiv.org/abs/2601.16163},
  doi_type = {arXiv preprint},
}

@inproceedings{li2025uva,
  title = {{Unified Video Action Model}},
  author = {Li, Shuang and Gao, Yihuai and Sadigh, Dorsa and Song, Shuran},
  year = {2025},
  eprint = {2503.00200},
  archivePrefix = {arXiv},
  doi = {10.15607/RSS.2025.XXI.074},
  url = {https://www.roboticsproceedings.org/rss21/p074.html},
  doi_type = {Published version},
  booktitle = {Proceedings of Robotics: Science and Systems},
}

@misc{lin2026jepawam,
  title = {{JEPA-WAM: Learning Vision-Language-Action Policies with Joint-Embedding World Modeling}},
  author = {Lin, Yihan and He, Jiawei and Bao, Shifeng and Zhao, Chen and Li, Yang and Wang, Xiaobo and Wang, Yan and Chi, Cheng and Zhang, Jing},
  year = {2026},
  eprint = {2608.09381},
  archivePrefix = {arXiv},
  doi = {10.48550/arXiv.2608.09381},
  url = {https://arxiv.org/abs/2608.09381},
  doi_type = {arXiv preprint},
}

@inproceedings{liu2026lara,
  title = {{LARA: Latent Action Representation Alignment for Vision-Language-Action Models}},
  author = {Liu, Mengya and Jia, Baoxiong and Huang, Jiangyong and Zhang, Jingze and Huang, Siyuan},
  year = {2026},
  eprint = {2606.07100},
  archivePrefix = {arXiv},
  doi = {10.48550/arXiv.2606.07100},
  url = {https://icml.cc/virtual/2026/poster/61232},
  doi_type = {arXiv preprint},
  booktitle = {Proceedings of the 43rd International Conference on Machine Learning},
}

@misc{liu2026leapbot,
  title = {{LeapBot-WA: World-Anchor Action Models via Predictive Latent Alignments}},
  author = {Liu, Pei and Zheng, Nan and Zhang, Lang and Peng, Daojie and Zhang, Yanan and Kong, Feilong and Feng, Mingyue and Liu, Jiachao and Wang, Yaonong and Chen, Qifeng and Ma, Jun},
  year = {2026},
  eprint = {2607.23969},
  archivePrefix = {arXiv},
  doi = {10.48550/arXiv.2607.23969},
  url = {https://arxiv.org/abs/2607.23969},
  doi_type = {arXiv preprint},
}

@misc{miao2026jepavla,
  title = {{JEPA-VLA: Video Predictive Embedding is Needed for VLA Models}},
  author = {Miao, Shangchen and Feng, Ningya and Wu, Jialong and Lin, Ye and He, Xu and Li, Dong and Long, Mingsheng},
  year = {2026},
  eprint = {2602.11832},
  archivePrefix = {arXiv},
  doi = {10.48550/arXiv.2602.11832},
  url = {https://arxiv.org/abs/2602.11832},
  doi_type = {arXiv preprint},
}

@inproceedings{octo2024policy,
  title = {{Octo: An Open-Source Generalist Robot Policy}},
  author = {Dibya Ghosh and Homer Rich Walke and Karl Pertsch and Kevin Black and Oier Mees and Sudeep Dasari and Joey Hejna and Tobias Kreiman and Charles Xu and Jianlan Luo and You Liang Tan and Lawrence Yunliang Chen and Quan Vuong and Ted Xiao and Pannag R Sanketi and Dorsa Sadigh and Chelsea Finn and Sergey Levine},
  booktitle = {Proceedings of Robotics: Science and Systems},
  year = {2024},
  doi = {10.15607/RSS.2024.XX.090},
  url = {https://www.roboticsproceedings.org/rss20/p090.html},
  doi_type = {Published version},
}

@inproceedings{reuss2024mdt,
  title = {{Multimodal Diffusion Transformer: Learning Versatile Behavior from Multimodal Goals}},
  author = {Moritz Reuss and \"{O}mer Erdin\c{c} Ya\u{g}murlu and Fabian Wenzel and Rudolf Lioutikov},
  booktitle = {Proceedings of Robotics: Science and Systems},
  year = {2024},
  doi = {10.15607/RSS.2024.XX.121},
  url = {https://www.roboticsproceedings.org/rss20/p121.html},
  doi_type = {Published version},
}

@inproceedings{sun2026vlajepa,
  title = {{VLA-JEPA: Enhancing Vision-Language-Action Model with Latent World Model}},
  author = {Sun, Jingwen and Zhang, Wenyao and Qi, Zekun and Ren, Shaojie and Liu, Zezhi and Zhu, Hanxin and Sun, Guangzhong and Jin, Xin and Chen, Zhibo},
  year = {2026},
  eprint = {2602.10098},
  archivePrefix = {arXiv},
  doi = {10.1007/978-3-032-37196-6_26},
  booktitle = {Computer Vision -- ECCV 2026},
  volume = {17064},
  pages = {478--497},
}

@misc{xu2026jepapolicy,
  title = {{JEPA Policy: Diffusion-Free Imitation Learning via Paired Action and Future Representation Prediction}},
  author = {Xu, Jie and Yu, Kangjin and Jin, Ziyi and Gao, Junjie and Chen, Liqing and Li, Yixian and Tian, Shuai and Xia, Zhongpu},
  year = {2026},
  eprint = {2609.09630},
  archivePrefix = {arXiv},
  doi = {10.48550/arXiv.2609.09630},
  url = {https://arxiv.org/abs/2609.09630},
  doi_type = {arXiv preprint},
}

@misc{ye2026dreamzero,
  title = {{World Action Models are Zero-shot Policies}},
  author = {Ye, Seonghyeon and Ge, Yunhao and Zheng, Kaiyuan and Gao, Shenyuan and Yu, Sihyun and Kurian, George and Indupuru, Suneel and Tan, You Liang and Zhu, Chuning and Xiang, Jiannan and Malik, Ayaan and Lee, Kyungmin and Liang, William and Ranawaka, Nadun and Gu, Jiasheng and Xu, Yinzhen and Wang, Guanzhi and Hu, Fengyuan and Narayan, Avnish and Bjorck, Johan and Wang, Jing and Kim, Gwanghyun and Niu, Dantong and Zheng, Ruijie and Xie, Yuqi and Wu, Jimmy and Wang, Qi and Julian, Ryan and Xu, Danfei and Du, Yilun and Chebotar, Yevgen and Reed, Scott and Kautz, Jan and Zhu, Yuke and Fan, Linxi "Jim" and Jang, Joel},
  year = {2026},
  eprint = {2602.15922},
  archivePrefix = {arXiv},
  doi = {10.48550/arXiv.2602.15922},
  url = {https://arxiv.org/abs/2602.15922},
}

@inproceedings{yu2025repa,
  title = {{Representation Alignment for Generation: Training Diffusion Transformers Is Easier Than You Think}},
  author = {Yu, Sihyun and Kwak, Sangkyung and Jang, Huiwon and Jeong, Jongheon and Huang, Jonathan and Shin, Jinwoo and Xie, Saining},
  booktitle = {International Conference on Learning Representations},
  year = {2025},
  pages = {87400--87442},
  eprint = {2410.06940},
  archivePrefix = {arXiv},
  doi = {10.48550/arXiv.2410.06940},
}

@misc{zhao2026frappe,
  title = {{FRAPPE: Infusing World Modeling into Generalist Policies via Multiple Future Representation Alignment}},
  author = {Zhao, Han and Wang, Jingbo and Song, Wenxuan and Chen, Shuai and Liu, Yang and Wang, Yan and Li, Haoang and Wang, Donglin},
  year = {2026},
  eprint = {2602.17259},
  archivePrefix = {arXiv},
  doi = {10.48550/arXiv.2602.17259},

}

@inproceedings{zheng2025flare,
  title = {{FLARE: Robot Learning with Implicit World Modeling}},
  author = {Ruijie Zheng and Jing Wang and Scott Reed and Johan Bjorck and Yu Fang and Fengyuan Hu and Joel Jang and Kaushil Kundalia and Zongyu Lin and Lo\"{i}c Magne and Avnish Narayan and You Liang Tan and Guanzhi Wang and Qi Wang and Jiannan Xiang and Yinzhen Xu and Seonghyeon Ye and Jan Kautz and Furong Huang and Yuke Zhu and Linxi Fan},
  booktitle = {Proceedings of The 9th Conference on Robot Learning},
  year = {2025},
  volume = {305},
  pages = {3952--3971},
  publisher = {PMLR},
  eprint = {2505.15659},
  archivePrefix = {arXiv},
  doi = {10.48550/arXiv.2505.15659},
  url = {https://proceedings.mlr.press/v305/zheng25a.html},
}

@inproceedings{zitkovich2023rt2,
  title = {{RT-2: Vision-Language-Action Models Transfer Web Knowledge to Robotic Control}},
  author = {Brianna Zitkovich and Tianhe Yu and Sichun Xu and Peng Xu and Ted Xiao and Fei Xia and Jialin Wu and Paul Wohlhart and Stefan Welker and Ayzaan Wahid and Quan Vuong and Vincent Vanhoucke and Huong Tran and Radu Soricut and Anikait Singh and Jaspiar Singh and Pierre Sermanet and Pannag R. Sanketi and Grecia Salazar and Michael S. Ryoo and Krista Reymann and Kanishka Rao and Karl Pertsch and Igor Mordatch and Henryk Michalewski and Yao Lu and Sergey Levine and Lisa Lee and Tsang-Wei Edward Lee and Isabel Leal and Yuheng Kuang and Dmitry Kalashnikov and Ryan Julian and Nikhil J. Joshi and Alex Irpan and Brian Ichter and Jasmine Hsu and Alexander Herzog and Karol Hausman and Keerthana Gopalakrishnan and Chuyuan Fu and Pete Florence and Chelsea Finn and Kumar Avinava Dubey and Danny Driess and Tianli Ding and Krzysztof Marcin Choromanski and Xi Chen and Yevgen Chebotar and Justice Carbajal and Noah Brown and Anthony Brohan and Montserrat Gonzalez Arenas and Kehang Han},
  booktitle = {Proceedings of The 7th Conference on Robot Learning},
  year = {2023},
  volume = {229},
  pages = {2165--2183},
  publisher = {PMLR},
  eprint = {2307.15818},
  archivePrefix = {arXiv},
  doi = {10.48550/arXiv.2307.15818},
  url = {https://proceedings.mlr.press/v229/zitkovich23a.html},
}

@inproceedings{black2025pi05,
  title = {{$\pi_{0.5}$: A Vision-Language-Action Model with Open-World Generalization}},
  author = {Kevin Black and Noah Brown and James Darpinian and Karan Dhabalia and Danny Driess and Adnan Esmail and Michael Robert Equi and Chelsea Finn and Niccolo Fusai and Manuel Y. Galliker and Dibya Ghosh and Lachy Groom and Karol Hausman and Brian Ichter and Szymon Jakubczak and Tim Jones and Liyiming Ke and Devin LeBlanc and Sergey Levine and Adrian Li-Bell and Mohith Mothukuri and Suraj Nair and Karl Pertsch and Allen Z. Ren and Lucy Xiaoyang Shi and Laura Smith and Jost Tobias Springenberg and Kyle Stachowicz and James Tanner and Quan Vuong and Homer Walke and Anna Walling and Haohuan Wang and Lili Yu and Ury Zhilinsky},
  booktitle = {Proceedings of The 9th Conference on Robot Learning},
  year = {2025},
  volume = {305},
  pages = {17--40},
  publisher = {PMLR},
  eprint = {2504.16054},
  archivePrefix = {arXiv},
  doi = {10.48550/arXiv.2504.16054},
  url = {https://proceedings.mlr.press/v305/black25a.html},
}

@article{oquab2023dinov2,
  title = {{DINOv2}: Learning Robust Visual Features without Supervision},
  author = {Oquab, Maxime and Darcet, Timoth\'ee and Moutakanni, Th\'eo and Vo, Huy and Szafraniec, Marc and Khalidov, Vasil and Fernandez, Pierre and Haziza, Daniel and Massa, Francisco and El-Nouby, Alaaeldin and Assran, Mahmoud and Ballas, Nicolas and Galuba, Wojciech and Howes, Russell and Huang, Po-Yao and Li, Shang-Wen and Misra, Ishan and Rabbat, Michael and Sharma, Vasu and Synnaeve, Gabriel and Xu, Hu and J\'egou, Herv\'e and Mairal, Julien and Labatut, Patrick and Joulin, Armand and Bojanowski, Piotr},
  journal = {arXiv preprint arXiv:2304.07193},
  year = {2023}
}

@inproceedings{fei2026liberoplus,
  author = {Fei, Senyu and Wang, Siyin and Shi, Junhao and Dai, Zihao and Cai, Jikun and Qian, Pengfang and Ji, Li and He, Xinzhe and Zhang, Shiduo and Fei, Zhaoye and Fu, Jinlan and Gong, Jingjing and Qiu, Xipeng},
  title = {{LIBERO-Plus}: A Progressive Robustness Benchmark for Visual-Language-Action Models},
  booktitle = {Proceedings of the IEEE/CVF Conference on Computer Vision and Pattern Recognition},
  year = {2026},
  pages = {38574--38583}
}
